\documentclass{bmvc2k}

\title{Positive Pair Distillation Considered Harmful:
Continual Meta Metric Learning for
Lifelong Object Re-Identification}

\addauthor{Kai Wang}{kwang@cvc.uab.es}{1*}
\addauthor{Chenshen Wu}{chenshen@cvc.uab.es}{1*}
\addauthor{Andrew D. Bagdanov}{andrew.bagdanov@unifi.it}{3}
\addauthor{Xialei Liu (\textit{corresponding author})}{xialei@nankai.edu.cn}{2}
\addauthor{Shiqi Yang}{syang@cvc.uab.es}{1}
\addauthor{Shangling Jui}{jui.shangling@huawei.com}{4}
\addauthor{Joost van de Weijer}{joost@cvc.uab.es}{1}

\addinstitution{
 Computer Vision Center\\
 Universitat Autònoma de Barcelona\\
 Barcelona, Spain
}
\addinstitution{
 College of Computer Science\\
 Nankai University\\
 Tianjin, China
}
\addinstitution{
 MICC \\
 University of Florence\\
 Florence, Italy
}
\addinstitution{
 Huawei Kirin Solution\\
 Shanghai, China
}

\runninghead{Kai Wang et al.}{Positive Pair Distillation Considered Harmful}

\usepackage{color}
\usepackage{graphicx}
\usepackage{booktabs}

\newcommand{\minisection}[1]{\vspace{0.01in} \noindent {\bf #1}\ }
\usepackage{amsmath}
\usepackage{amssymb}
\usepackage{multirow}
\usepackage{pifont}
\usepackage{enumitem}
\usepackage{wrapfig,lipsum,booktabs}
\usepackage{wrapfig}

\usepackage{xcolor}

\usepackage[final]{pdfpages}

\begin{document}

\maketitle
\def\thefootnote{*}\footnotetext{These authors contributed equally to this work.}

\begin{abstract}
Lifelong object re-identification incrementally learns from a stream of re-identification tasks. The objective is to learn a representation that can be applied to all tasks and that generalizes to previously \textit{unseen} re-identification tasks. The main challenge is that at inference time the representation must generalize to previously \textit{unseen} identities. To address this problem, we apply continual meta metric learning to lifelong object re-identification. To prevent forgetting of previous tasks, we use knowledge distillation and explore the roles of positive and negative pairs. Based on our observation that the distillation and metric losses are \emph{antagonistic}, we propose to remove positive pairs from distillation to robustify model updates. Our method, called Distillation without Positive Pairs (DwoPP), is evaluated on extensive intra-domain experiments on person and vehicle re-identification datasets, as well as inter-domain experiments on the LReID benchmark. Our experiments demonstrate that DwoPP significantly outperforms the state-of-the-art.
\end{abstract}

\section{Introduction}
\label{sec:intro}

\begin{figure}[t]
\centerline{\includegraphics[width=0.67\linewidth]{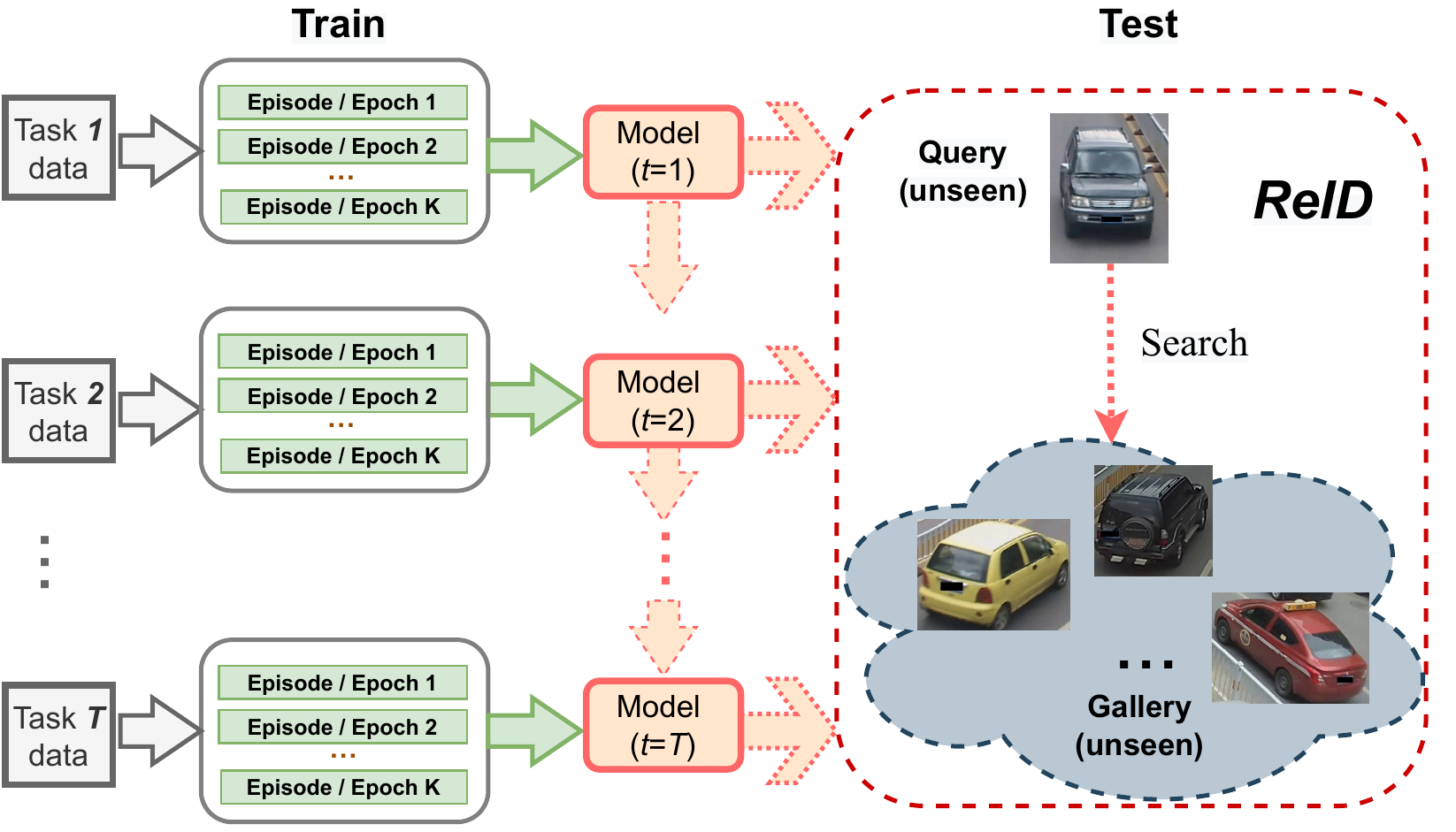}}
\caption{Lifelong Object ReID with continual meta-metric learning. Unlike conventional object re-identification, data are presented sequentially in discrete tasks of disjoint classes. Data from previous tasks are unavailable in successive ones and the learner must incrementally update when a new task arrives. Furthermore, in object re-identification the test identities are not seen during training, which demands  generalization of the learned metric.}
\label{fig:cml_setup}
\end{figure}

\emph{Object re-identification (ReID)} aims to associate the identity of a query image with those in a gallery set~\cite{he2021transreid,zhang2021refining}. 
It is applied to many applications, including person re-identification~\cite{chen2019spatial,Li_2021_CVPR,ye2021deep}, vehicle re-identification~\cite{Khorramshahi_2019_ICCV,lou2019embedding,zhao2021phd}, and face verification~\cite{wang2018additive,wang2017normface}. 
Most existing approaches assume that the test and training dataset are drawn from the same distribution and that all training data is available jointly when training the network~\cite{Li_2021_CVPR,lou2019embedding,Luo_2019_CVPR_Workshops,zhang2021refining,zhao2021phd}. 
In domain generalization ReID~\cite{ni2022meta,song2019generalizable,choi2021meta,bai2021person30k,dai2021generalizable} all source domain data is assumed available during training.
This assumption is not realistic for many applications as all training data might not be available from the start and its distribution could vary over time. In addition, the trained system could be applied at inference time to new data never seen during training. 
Only recently, the problem of Lifelong ReID has been proposed~\cite{pu2021lifelong}. This setting requires learning from a \textit{sequence} of domains, and evaluates the algorithm on \textit{unseen} domains.


Continual learning~\cite{mccloskey1989catastrophic,de2019continual,masana2020class,lomonaco2021avalanche} addresses the problem of learning from non-stationary streams of data. 
It has developed several techniques including regularization-based methods~\cite{aljundi2018memory,gomez2022continually,kirkpatrick2017overcoming,liu2018rotate,zenke2017continual}, parameter-isolation~\cite{mallya2018piggyback,masana2021ternary,mallya2018packnet,serra2018overcoming}, and replay-based methods~\cite{hayes2019remind,liu2020generative,wu2018memory,wu2019large,wang2021acae,yan2021dynamically}. In this paper we consider exemplar-free continual learning where it is not allowed to save any samples (exemplars) of previous tasks for the problem of object re-identification. 
This requirement is out of the privacy considerations in person ReID problems.

Most continual learning methods specifically consider the incremental learning of classification problems. The considered setup for object re-identification (Fig.~\ref{fig:cml_setup}) is different in two main aspects. Firstly, they usually do not incrementally learn a \textit{classifier}, instead they incrementally learn a feature representation. 
Secondly, the aim is to perform evaluation on new unseen tasks. So the real goal is to incrementally learn a metric space that generalizes to previously unseen tasks. Pu et al.~\cite{pu2021lifelong} propose a method to address the first problem but ignore the second consideration: \textit{the representation should generalize to unseen tasks}.

Meta-learning~\cite{Baik_2021_ICCV,Chen_2021_ICCV,finn2017model,hospedales2020meta,nichol2018first,snell2017prototypical,vilalta2002perspective} focus on generalising to unseen tasks and has been applied to few-shot learning~\cite{bateni2020improved,li2020adversarial,liu2019prototype,su2020does,wang2020generalizing,yang2021free}.
Object ReID can be considered a few-shot learning problem,  since the object identities at test time are not shown during the training and we only have few  support images. 
To exploit the generalization capability of meta learning, Chen et al.~\cite{chen2019deep} propose Deep Meta Metric Learning (DMML) that formulates the deep metric learning as a meta learning problem. 
Since the main challenges of object re-identification are learning from a sequences and generalization to previously unseen domains, in this paper we propose \emph{Continual Meta Metric Learning} to address this problem.

To further endow continual meta metric learning with a mechanism to mitigate forgetting knowledge from previous tasks, we introduce a temporary classifier for the support set and study the potential of directly applying knowledge distillation~\cite{hinton2015distilling,li2017learning}.
However, we find that the distillation and metric learning losses are antagonistic.
We therefore propose \textbf{D}istillation \textbf{w}ith\textbf{o}ut \textbf{P}ositive \textbf{P}airs (DwoPP). DwoPP, different from naive distillation, which distills knowledge from the previous to the current task classifier over \textit{all} classes in the current task, distills only using \textit{negative} examples. In this way, we avoid the antagonistic relationship between the metric and distillation losses which is from positive pairs distillation. 

The main contributions of our paper are: \textbf{1)} we show that meta metric learning is superior to global metric learning for object re-identification; \textbf{2)} we explicitly explore the roles of positive and negative pairs in distillation and propose a novel distillation scheme called DwoPP for Continual Meta Metric Learning; \textbf{3)} we propose task splits for evaluation of continual metric learning methods on intra-domain object ReID for three ReID datasets and evaluate on much longer sequences than existing benchmarks; and \textbf{4)} we perform extensive experimental analysis demonstrating that, DwoPP achieves significantly better performance on person and vehicle ReID, as well as on the lifelong re-identification (LReID) benchmark~\cite{pu2021lifelong}.

\section{Related work}
\label{sec:related_work}
\minisection{Object re-identification and metric learning.}
Metric learning has been widely applied to object re-identification~\cite{he2021transreid,zhang2021refining}, mainly focusing on person ReID~\cite{chen2019spatial,Li_2021_CVPR,ye2021deep,zhao2021continual}, vehicle ReID~\cite{Khorramshahi_2019_ICCV,lou2019embedding,zhao2021phd} and face verification~\cite{wang2018additive,wang2017normface,zhao2021continual}).
Deep metric learning methods can be divided into three categories based on the loss used: contrastive loss with pairwise inputs~\cite{chopra2005learning}, triplet loss with triplet inputs~\cite{hoffer2015deep}, and N-pair loss with batch inputs~\cite{simo2015discriminative}.
In general, deep metric learning works well but does not take generalization of the learned metrics into account and neglects relationships between inter-class samples.  DMML~\cite{chen2019deep} formulates metric learning for object re-identification from a meta learning perspective.
We build upon DMML for our \textit{continual learning} view of meta metric learning.


\minisection{Continual learning.}
\label{subsec:continual_learning}
Continual learning methods can be categorized into three groups: parameter-isolation, regularization-based and replay-based methods~\cite{de2019continual}. 
The most relevant to our work are regularization-based methods~\cite{yu2020semantic,aljundi2018memory,kirkpatrick2017overcoming,li2017learning,liu2018rotate,zenke2017continual,pelosin2022towards,liu2018rotate}.  
Knowledge distillation is a widely used regularization method which decreases forgetting by either aligning features~\cite{liu2020generative,wu2018memory} or the predicted probabilities~\cite{li2017learning}. 
To adapt knowledge distillation to Continual Meta Metric Learning, we propose a variant of knowledge distillation by introducing a temporary classifier for the current support set, and more importantly the distillation in the paper is  without considering positive pairs. Replay-based continual learning overcomes forgetting by saving a set of exemplars from each task~\cite{hou2019learning,wu2019large,hayes2019remind,hou2019learning,wu2018memory,wu2019large,yan2021dynamically,wang2022incremental}. We focus on exemplar-free continual learning. And continual learning applied to persons in particular has  privacy considerations which  makes retaining  data problematic.

\minisection{(Incremental) Meta learning.}
Meta learning based on metrics or optimization-based approaches are the main directions of current research~\cite{wang2020generalizing}. 
ProtoNets~\cite{snell2017prototypical} and RelationNets~\cite{sung2018learning} are canonical representatives of metric-based approaches, while MAML~\cite{finn2017model} and Reptile~\cite{nichol2018first} are representative optimization-based methods.
Incremental meta learning (IDA~\cite{liu2020incremental}, ERD~\cite{wang2021incremental}) methods have been mainly developed for incremental few-shot learning, however, they can also be applied to lifelong object ReID and we will compare to them in the experimental section.
There are a few methods on incremental metric learning which approach the problem as one of representation learning with a metric-based classification loss. Examples include CRL~\cite{zhao2021continual}, FGIR~\cite{chen2020exploration}, and AKA~\cite{pu2021lifelong}. However, these works all focus on distillation over seen classes and thus neglect the need to recognize unseen identities.

\section{Methodology}
\label{sec:method}


\subsection{Preliminaries}
There are two main approaches to metric learning applied to object ReID: those based on global optimization of a metric embedding over the training set, and those based on episodic meta learning.
Most global optimization metric learning methods minimize a metric loss over the whole dataset $\mathit{D}=(\mathbf{X},\mathbf{Y})$ of inputs $\mathbf{X}$ and corresponding labels $\mathbf{Y}$. For comparison in this paper we use the popular softmax-triplet loss as used in Bag-of-Tricks (BoT)~\cite{Luo_2019_CVPR_Workshops}.  

\minisection{Deep meta metric learning (DMML).} In DMML~\cite{chen2019deep}, the authors instead formulate metric learning as a meta learning problem. They decompose the training data into a series of sub-tasks, called \textit{episodes} in meta learning, and then learn a meta metric that generalizes well to all sub-tasks. Assuming the unseen test task is drawn from the same distribution of sub-tasks from the training set, this learned meta metric should generalize to this unseen test task. 

Assume that we sample $K$ episodes in total for training, that each episode $E_k$ is composed of $N$ classes, and that each class contains  $n_s$ images in the support set $\mathit{S}_k$ and $n_q$ images in the query set $\mathit{Q}_k$. In each episode, we learn the meta metric to correctly predict the query samples from support samples. The learning problem for DMML is:
\begin{equation}
  \label{eq:dmml_loss}
  \theta^{*}=\arg \min_{\theta} \mathbb{E}_{k \in [1,K]} \left[ \mathcal{L}_{\text{eps}}(\theta; \mathit{S}_k,\mathit{Q}_k) \right]
\end{equation}
where $\mathcal{L}_{\text{eps}}$ is the episode level hard-mining metric loss proposed in DMML~\cite{chen2019deep}. 

The episodic loss $\mathcal{L}_{\text{eps}}$ is defined in terms of positive and negative pairs. In the current episode $E_k$ with the class set $\mathbb{C}$, a query point $q_c \in Q_k$ is drawn from a specific class $c \in \mathbb{C}$.
We construct the \emph{positive pairs} $[q_c, s_c]$ from the query point and support points $s_c \in S_k$ from the class $c$, and negative pairs $[q_c,s_{c'}]$ from the query point and support points $s_{c'} \in S_k$ from \emph{different} classes $c' \neq c$. Hard mining is performed over the positive pairs by finding largest Euclidean distance from $q_c$ to a positive support sample $d_c= \max_{s_c \in S_k} d(q_c,s_c)$, and over negative pairs by finding the smallest distance from $q_c$ to the a negative support sample $d_{c'}= \min_{s_{c'} \in S_k} d(q_c,s_{c'})$. $\mathcal{L}_{\text{eps}}$ is defined in terms of these hard-mined distances ($\tau$ is a margin):
\begin{equation}
  \label{eq:eps_loss}
  \mathcal{L}_{\text{eps}}(\theta; \mathit{S}_k,\mathit{Q}_k) = \sum_{q_c \in Q_k} \log(1+ \!\!\! \!\!\! \sum_{c' \in \mathbb{C} \setminus \{c\}} \!\!\! \!\!\! \exp(d_{c'}-d_c+\tau)),
\end{equation}
\subsection{Continual Metric Learning}
\label{sec:cml}
In continual metric learning, tasks $ t \in [1,T]$ arrive sequentially as disjoint datasets $\mathit{D}_t$. The aim is to learn $\theta_t$ incrementally in a \emph{training session} for each task $t$ and to ensure it accumulates knowledge from the previous tasks so as to generalize better to unseen test tasks:
\begin{equation}
  \label{eq:cmml_loss}
  \theta^{*}_t=\arg \min_{\theta_t}  \mathcal{L}_{\text{cml}}(\theta_t; \mathit{D}_t).
\end{equation}
And the data from previous tasks (i.e. $D_{t'} \mbox{ for } t' < t$) are \textit{not} available to the learner at task $t$. Eq.~\ref{eq:cmml_loss} defines the \emph{continual} learning setup where one only has access to data of a single task at a time. It is a general equation applicable to continual learning setups. In Eq.~\ref{eq:cmml_loss} $\mathcal{L}_{\text{cml}}$ could be replaced with a metric learning loss (yielding continual metric learning) or with a meta metric learning loss, like Eq.~\ref{eq:dmml_loss}, to obtain continual meta metric learning. The challenge of continual learning is preventing forgetting of previous knowledge. 


\begin{figure*}[t]
\begin{center}
\begin{tabular}{cc}
\includegraphics[width=0.4\textwidth]{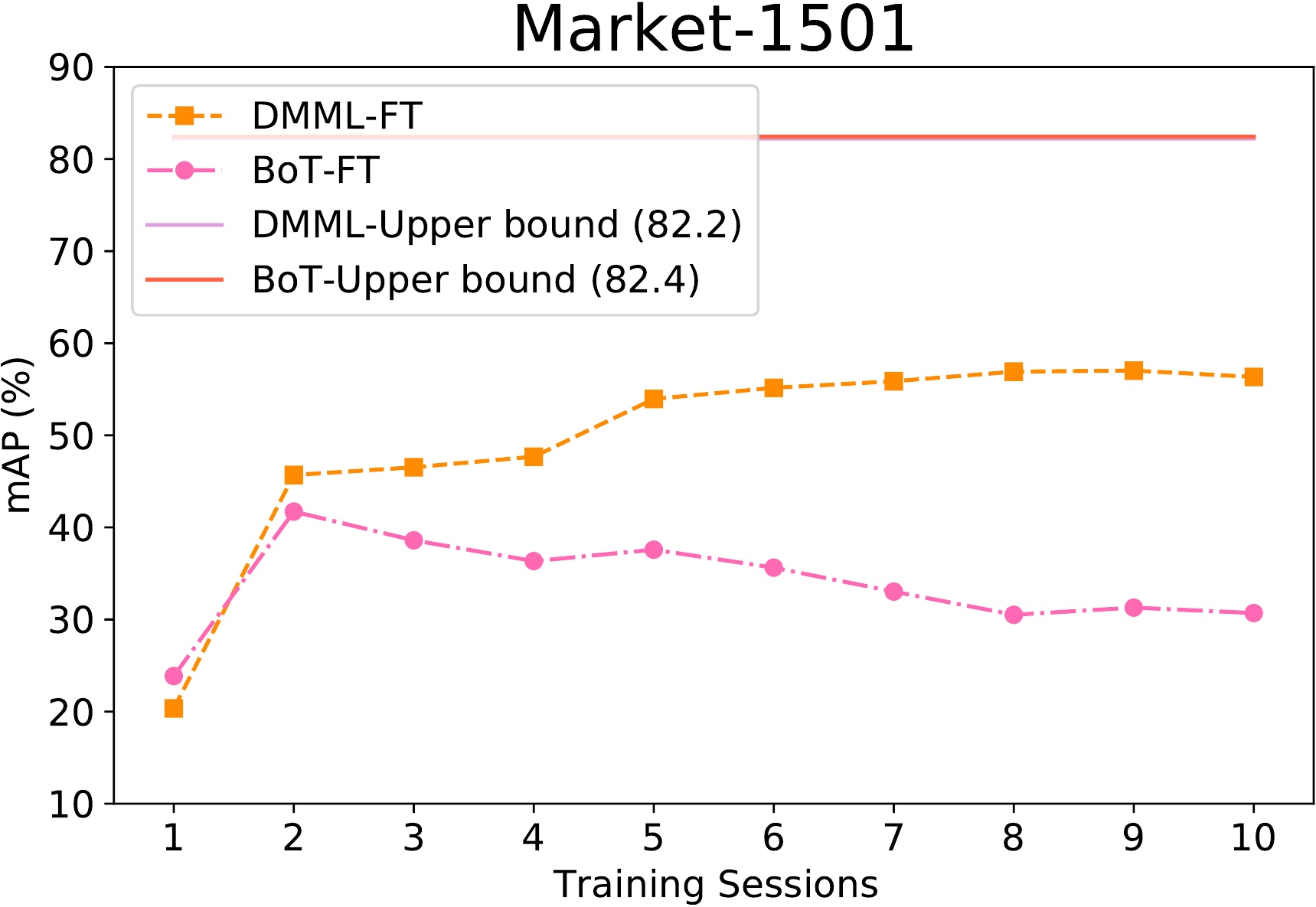} &
\includegraphics[width=0.4\textwidth]{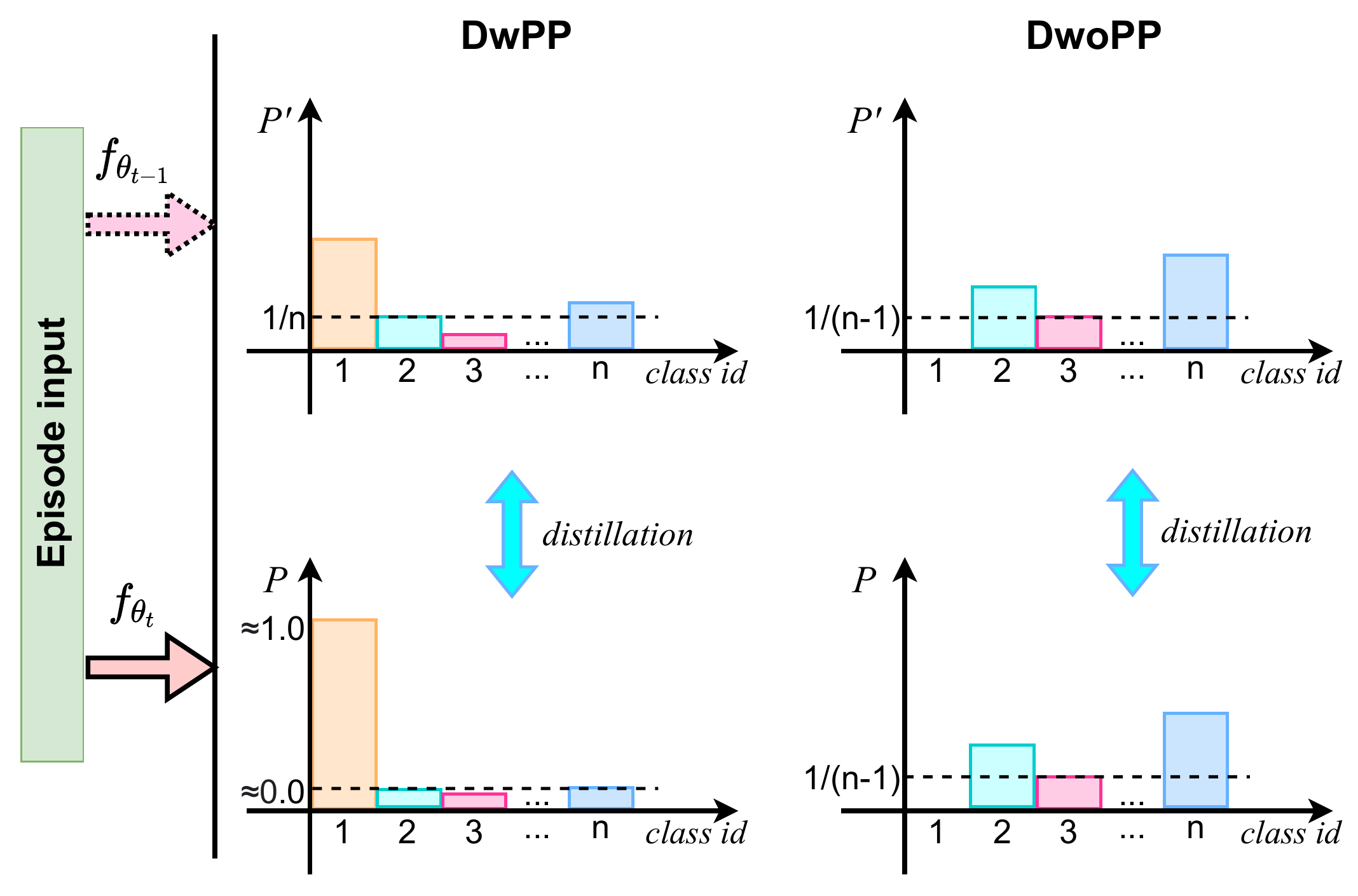} \\
(a) & (b)
\end{tabular}
\end{center}
\label{fig:simple_motivation}
\caption{(a) Comparing continual meta-metric learning  (DMML-FT~\cite{chen2019deep}) with continual metric learning (BoT-FT~\cite{Luo_2019_CVPR_Workshops}).
We finetune on 10 equally split Market-1501 tasks. Upper bounds are joint training on all data. (b) Comparison between DwPP and DwoPP (class 1 is the positive class). The old model has never seen class 1 and so likely produces an output less than 1 although we want positive pairs to map to the exact same point in latent space. Also, the dominance of the positive class inhibits distillation of negative pair information.
}
\label{fig:merge_fig2_fig3}
\end{figure*}


Here we consider two different losses for $\mathcal{L}_{cml}$ in Eq.~\ref{eq:cmml_loss}, either based on  meta-learning (like DMML~\cite{chen2019deep}) or on the softmax (like BoT~\cite{Luo_2019_CVPR_Workshops}). 
The majority of Person Re-Identification approaches (including the LReID benchmark~\cite{pu2021lifelong}) are based on the softmax-triplet loss. We compare these methods in the Continual Metric Learning setting on Market-1501 in Fig.~\ref{fig:merge_fig2_fig3}(a) by simply applying fine-tuning (FT) without any mitigation of forgetting. We clearly see that continual metric learning is quickly surpassed by continual meta-metric learning.
The underlying reason for this marked improvement is that re-identification aims to recognize \textit{unseen} objects (each object identity is represented by only one query image at test time). This is the central characteristic of few-shot recognition. 
Instead, the conventional softmax-triplet loss optimizes recognition on seen classes. It does not explicitly aim for generalization to unseen classes. Moreover, the focus on current task classes can also lead to increased forgetting of previous classes. The meta learning DMML loss, however, tends to learn a better representation space that generalizes to future unseen tasks and thus suffers less from forgetting. In brief, DMML is a more principled approach for continual metric learning than the softmax-triplet loss and we propose to use DMML as the basis.

\subsection{Distillation without Positive Pairs (DwoPP)}
\label{sec:DwoPP}

To adapt the DMML loss defined in Eq.~\ref{eq:eps_loss} to continual metric learning, we compute it for task $t$ over episodes $E_k^t$ drawn only from the current task data $D_t$. We denote the support set and query set of each episode during task $t$ as $S^t_k$ and $Q^t_k$. Then the DMML loss is defined with the current model $f_{\theta_t}$ as $\mathcal{L}_{\text{eps}}(\theta_t; S_k^t,Q_k^t)$ (see Eq.~\ref{eq:eps_loss}).

Episodic meta learning with the DMML loss will not mitigate forgetting in a continual metric learning. Knowledge Distillation~\cite{hinton2015distilling,li2017learning} is a common technique for alleviating catastrophic forgetting when learning over a sequence of tasks. Note, however, distillation assumes that a classifier over classes from the previous tasks is available on which to perform knowledge distillation -- something that for continual meta metric learning we do not have. However, based on the sampled episodes we can construct two temporary classifiers, one based on the previous and one based on the current tasks' feature extractor. We can then define a new distillation loss in terms of these temporary classifiers.

\minisection{Class Prototypes.} To construct the temporary classifier, we compute prototypes as the centroid of embedded samples of each class $\mathbf{u}_c$ ($c$ is class label): 
\begin{equation}
\label{eq:center}
    \mathbf{u}_c =\frac{1}{n_s}   \!\!\!\! \!\!\!\! \! \sum_{(x_i,y_i) \in S^t_{k}} \!\! f_{\theta}(x_i)\delta_c(y_i),
\end{equation}
where $\delta_c(y) = 1 \Leftrightarrow y=c$ is an indicator function. 

\minisection{DwPP: Distillation with Positive Pairs.} 
With the class prototypes $\mathbf{u}_c$, the prediction for class $c\in \mathbb{C}$ of query image $\hat x \in Q^t_k$ with the model $f_{\theta_t}$ is given by: 
\begin{align}
\label{eq:DwPP_classifier}
  g_{c}(S^t_{k},\hat{x};\theta) =  \frac{ [\exp(-d(f_{\theta}(\hat{x}),\mathbf{u}_{c}))]^{1/T} }{\sum_{c' \in \mathbb{C}}
       [\exp(-d(f_{\theta}(\hat{x}),\mathbf{u}_{c'}))]^{1/T} },
\end{align}
where $T$ is the temperature and $d$ is the Euclidean distance. These predictions are used to distill knowledge from task $t\text{-}1$ into  task $t$ by constructing two temporary classifiers, one using $\theta_t$ and another using $\theta_{t-1}$, and considering all negative and positive pairs: 
\begin{equation}
\label{eq:DwPP_distillation}
    \mathcal{L}_{\text{DwPP}} (\theta_{t}; \theta_{t-1},S^t_{k},Q^t_{k}) = 
    \sum_{ \hat{x} \in Q^t_{k} }  \!\!\! KL\left[  \mathbf{g}(S^t_{k}, \hat{x}; \theta_{t-1}) \, || \,  \mathbf{g}(S^t_{k}, \hat{x}; \theta_t)\right].
\end{equation}
Here $\mathbf{g}$ is a classifier constructed by concatenating the predictions $g_c$ defined in Eq.~\ref{eq:DwPP_classifier} for all classes in the episode. 

Knowledge distillation for continual meta metric learning requires careful attention to which pairs are included in the distillation loss. Consider the hypothetical case illustrated in Fig.~\ref{fig:merge_fig2_fig3}(b) where we show the predictions of the two temporary classifiers (class 1 is the query class). In task $t$, the new classes from $D_t$ are not well-discriminated from each other -- that is, the margin between positive and negative pairs in $D_t$ is not guaranteed by the model from task $t-1$ and the predicted probabilities are distributed as in the upper left column of Fig.~\ref{fig:merge_fig2_fig3}(b). After learning task $t$ we would like it to be a peaked distribution around the correct class, and simultaneously we also wish to maintain the relative probabilities of all classes (via knowledge distillation). Although this distillation will maintain model stability and mitigate forgetting, the estimate of the old model for the correct label is likely to be unreliable and will prevent the metric loss from pushing similar labels to the same position in the embedding space. Furthermore, the dominance of the positive class prevents distillation of the relevant negative pair information (also known as dark knowledge~\cite{hinton2015distilling}), which weakens the alignment of classes in the feature space.

In essence, the metric and distillation losses are \textit{antagonistic} due to the inclusion of positive pairs in knowledge distillation.
Thus we propose to remove positive pairs from distillation. As shown in the right column of Fig.~\ref{fig:merge_fig2_fig3}(b), since the other classes are negatives for class 1, they can be easily aligned with the previous probabilities to overcome forgetting. At the same time, the peaked distribution in the bottom left of Fig.~\ref{fig:merge_fig2_fig3}(b) can also be achieved by the metric loss. 
To further analyze the role of positive and negative pairs, we decouple the KL divergence into positive and negative pair distillation as proposed by DKD~\cite{zhao2022decoupled}, showing that positive pair distillation leads to performance degradation (see Table~\ref{tab:decoupled_kd}).

\minisection{DwoPP: Distillation without Positive Pairs.}
To remove positive pairs from DwPP distillation, we exclude class $\hat{y}$ which is the class label of the query image $\hat x \in Q^t_k$ from the temporary classifier and rewrite the Eq.~\ref{eq:DwPP_classifier} as:
\begin{align}
\label{eq:DwoPP_classifier}
g'_{c}(S^t_{k},\hat{x}, \hat{y};\theta) =  \frac{ [\exp(-d(f_{\theta}(\hat{x}),\mathbf{u}_{c}))]^{1/T} }{\sum_{c' \in \mathbb{C} \setminus \{\hat{y}\}}
       [\exp(-d(f_{\theta}(\hat{x}),\mathbf{u}_{c'}))]^{1/T} }
\end{align}
Then the DwoPP distillation can be rewritten as:
\begin{equation}
\label{eq:DwoPP_distillation}
    \mathcal{L}_{\text{DwoPP}} (\theta_{t}; \theta_{t-1},S^t_{k},Q^t_{k}) = \!\!\!\!\!
    \sum_{ (\hat{x}, \hat{y}) \in Q^t_{k} }  \!\!\!\!\! KL[  \mathbf{g}'(S^t_{k}, \hat{x}, \hat{y}; \theta_{t-1}) \, || \,  \mathbf{g}'(S^t_{k}, \hat{x}, \hat{y}; \theta_t)].
\end{equation}
With the above defined DwoPP distillation loss and episode DMML loss, the continual metric learning loss function for each episode is defined as:
\begin{equation}
\label{eq:final_loss}
\mathcal{L}_{\text{cml}}(\theta_t; \theta_{t-1}, \mathit{S}_k^t,\mathit{Q}_k^t) =   \mathcal{L}_{\text{eps}}(\theta_t; \mathit{S}_k^t,\mathit{Q}_k^t)  +  \lambda \mathcal{L}_{\text{DwoPP}}  (\theta_{t};\theta_{t-1},S^t_{k},Q^t_{k}).   
\end{equation}
To demonstrate the necessity of removing positive pairs from the distillation, we compare DwPP and DwoPP in Sec.~\ref{sec:experiments} and perform an ablation on $T$ in both. 

\begin{figure*}
\setlength{\tabcolsep}{0pt}

\begin{center}
\begin{tabular}{ccc}
\includegraphics[width=0.3\textwidth]{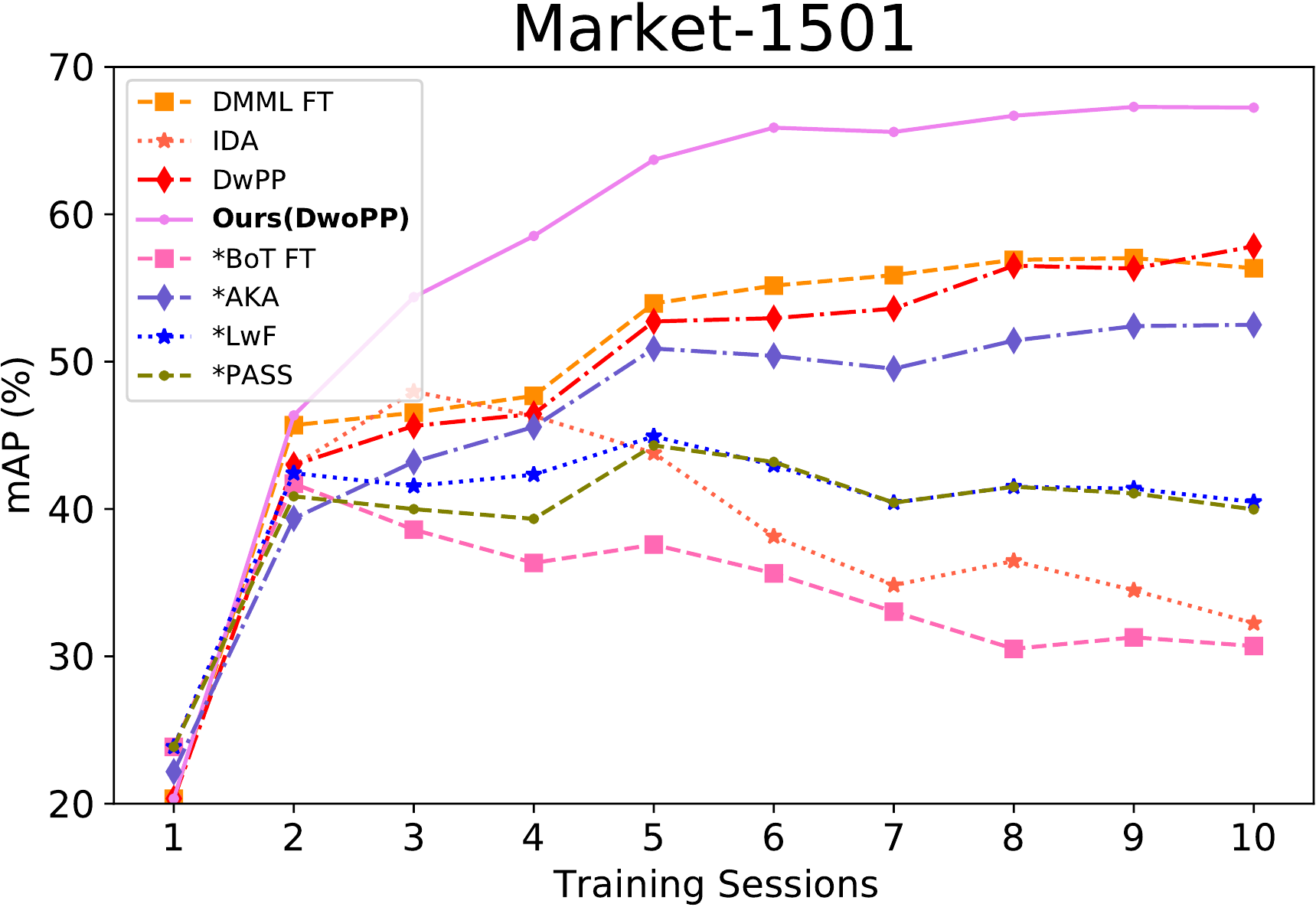} &
\includegraphics[width=0.3\textwidth]{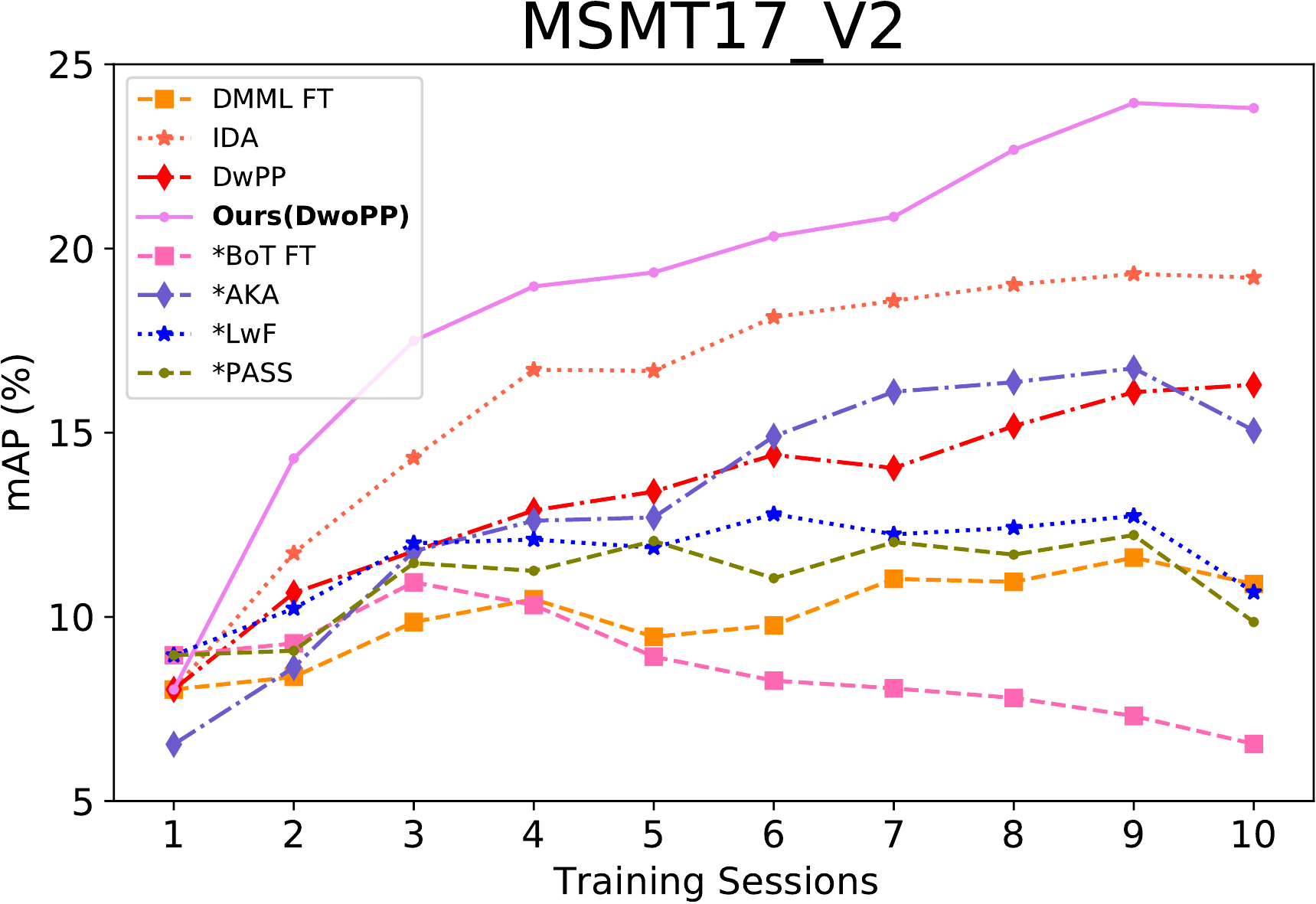} &
\includegraphics[width=0.3\textwidth]{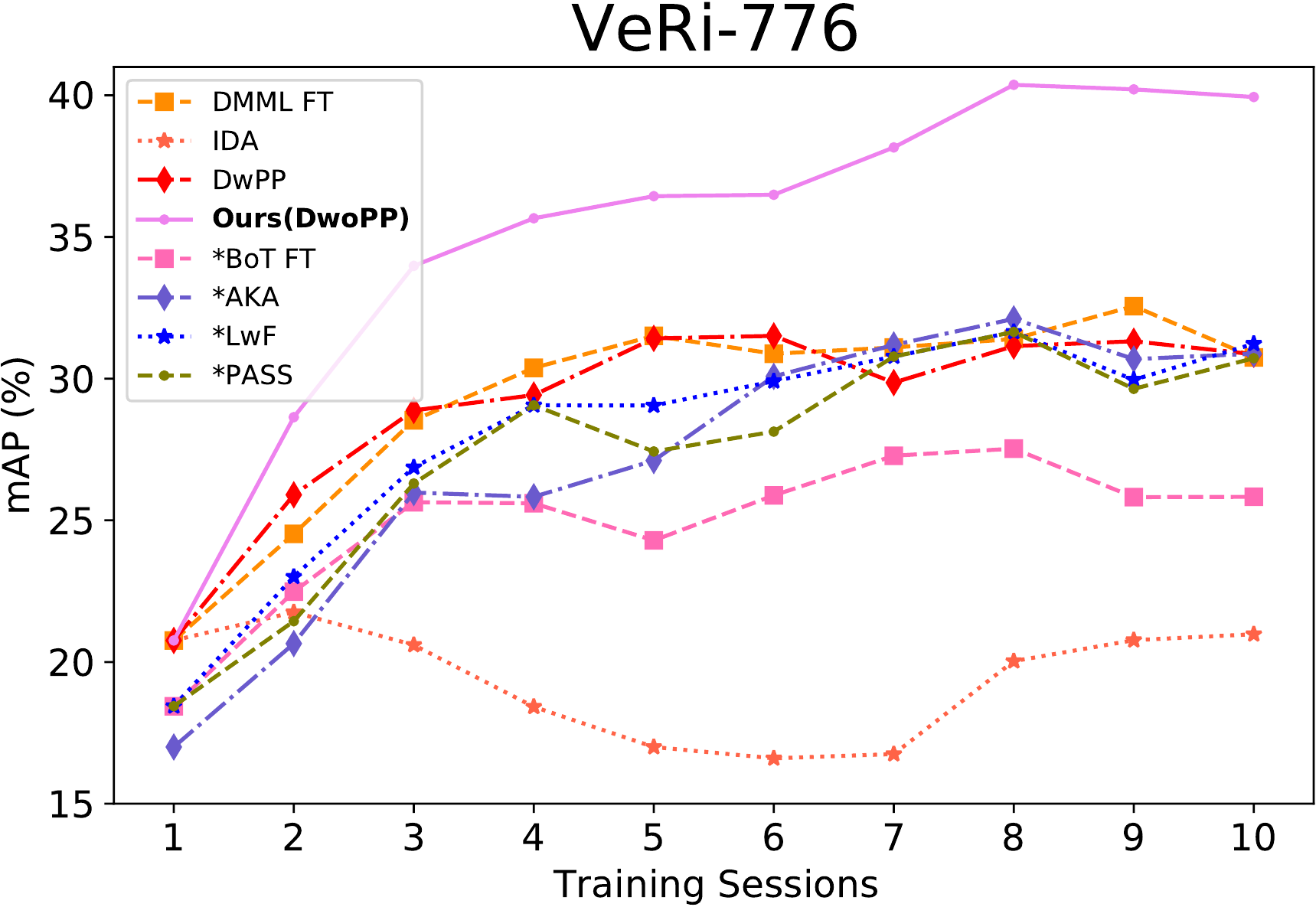} \\
(a) mAP on Market-1501 & (b) mAP on MSMT17\_V2 & (c) mAP on VeRi-776 \\

\includegraphics[width=0.3\textwidth]{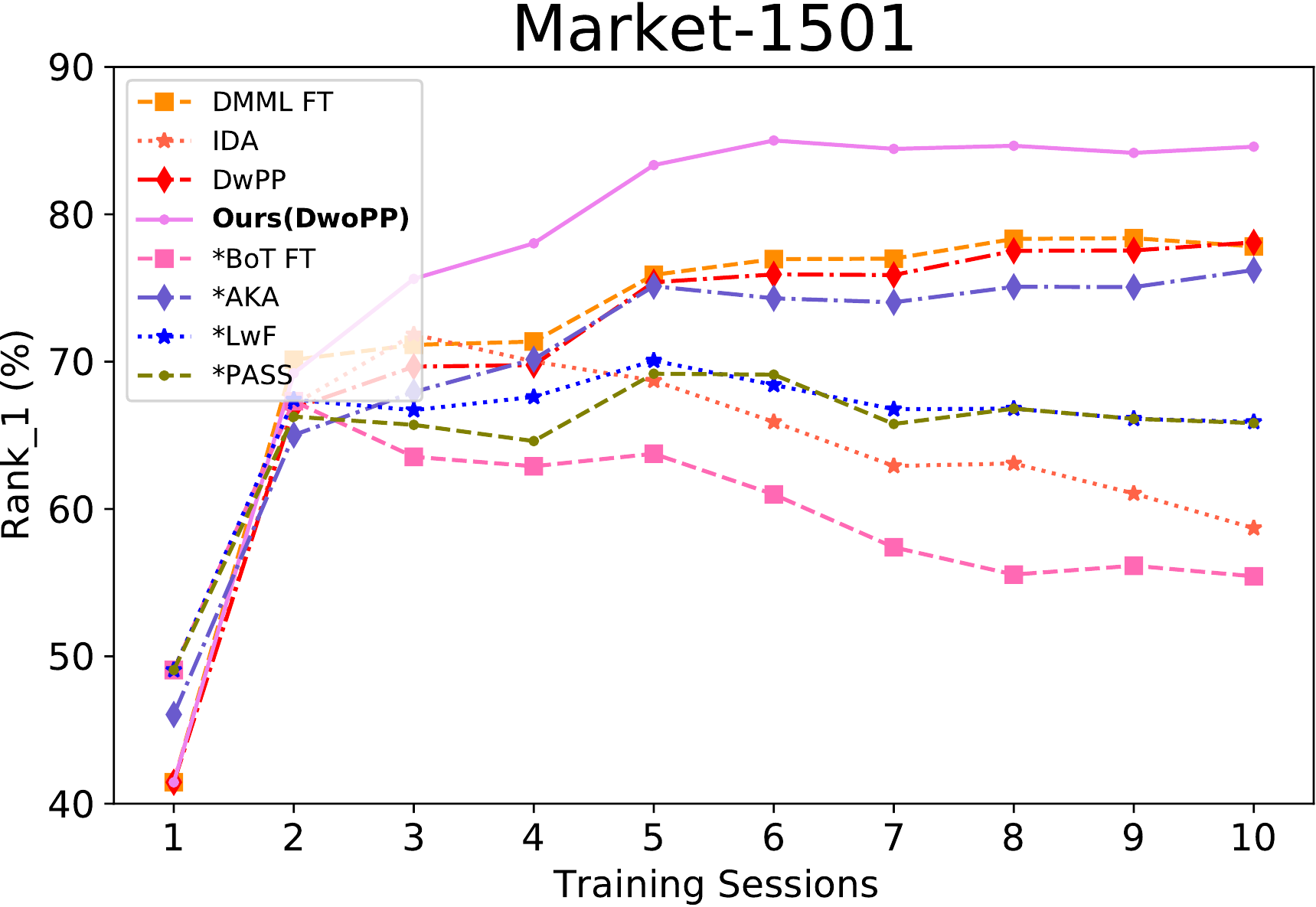} &
\includegraphics[width=0.3\textwidth]{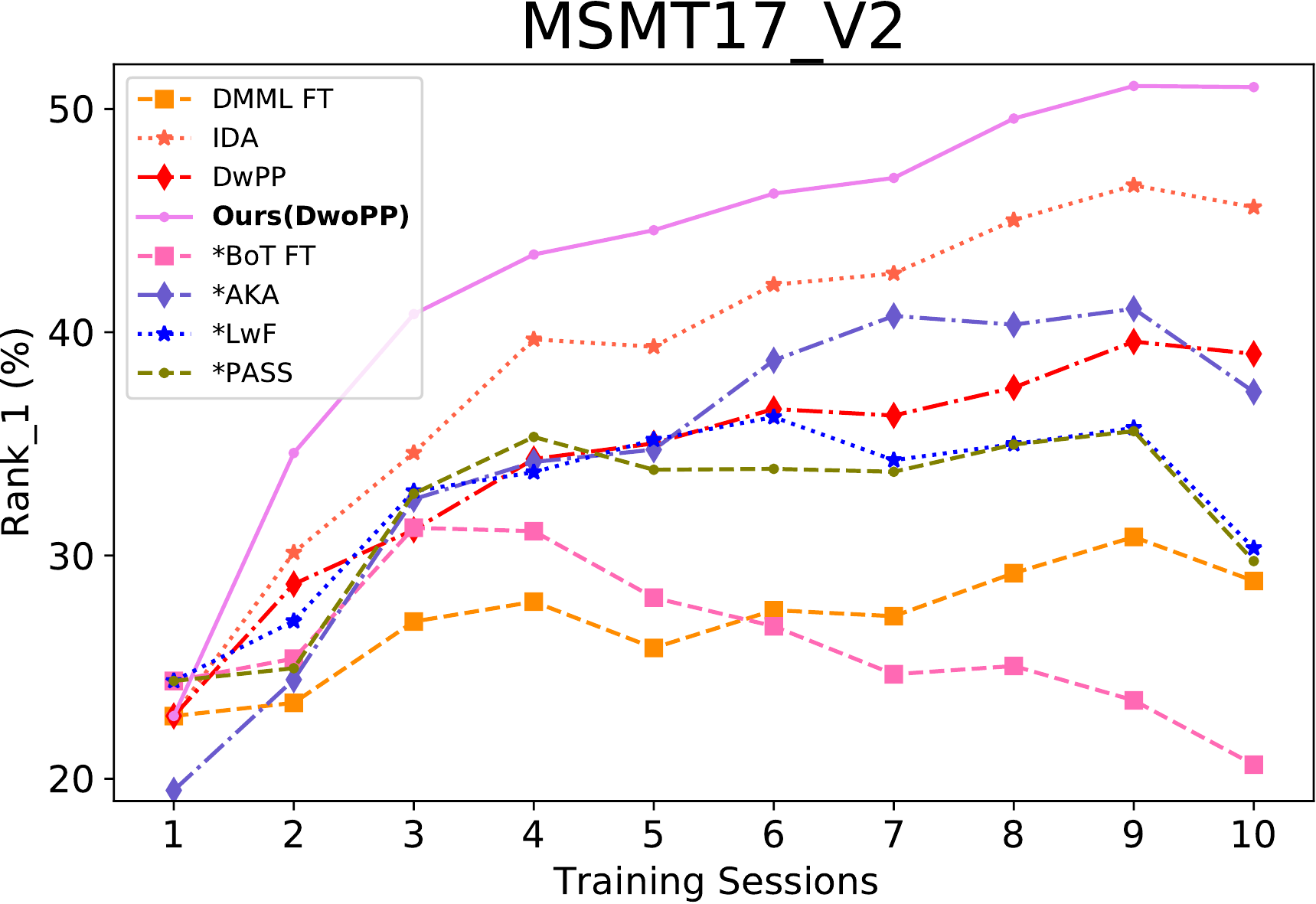} &
\includegraphics[width=0.3\textwidth]{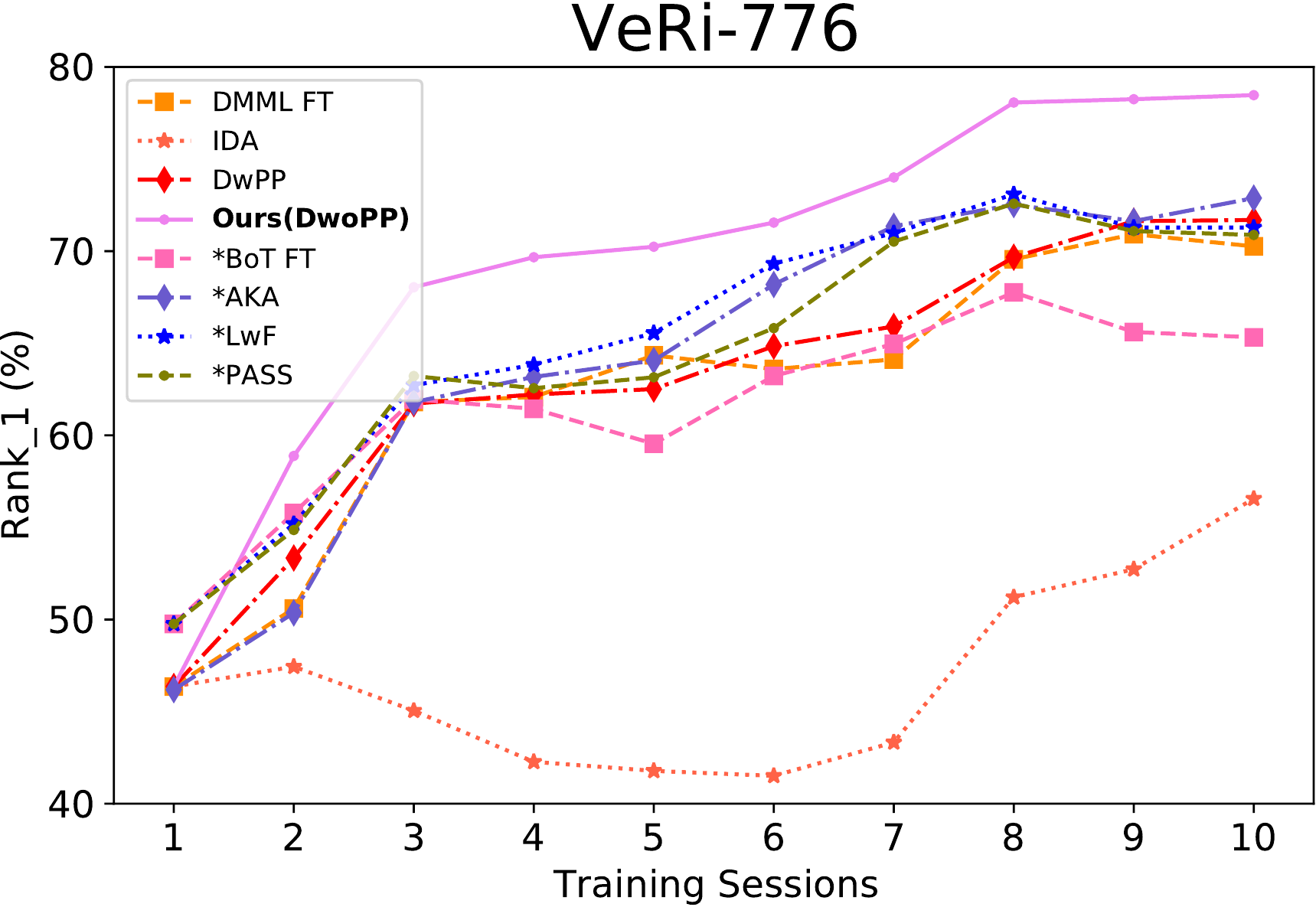} \\
(d) Rank-1 on Market-1501 & (e) Rank-1 on MSMT17\_V2 & (f) Rank-1 on VeRi-776
\end{tabular}
\end{center}
\caption{mAP and Rank-1 performance. Methods with ``*'' use the softmax-triplet loss.
}
\label{fig:10_task_all_dataset}
\end{figure*}

\section{Experimental Results}

\label{sec:experiments}

\subsection{Experimental setup}


\minisection{Datasets for Intra-domain Object ReID.} We propose continual metric learning splits for two Person ReID datasets and one vehicle ReID dataset.
\textbf{(1) Market-1501~\cite{zheng2015scalable}} consists of 32,668 images of 1,501 identities captured by 6 cameras. The dataset is divided into a training set with 12,968 images of 751 identities and a test set containing 3,368 query images and 19,732 gallery images of 750 identities. For continual metric learning setup, we split the 751 training identities into 10 disjoint tasks, each with 75 identities (the first with 76). 
\textbf{(2) MSMT17\_V2~\cite{wei2018person}} consists of 126,441 images of 4101 persons captured by 15 cameras. Its training set includes 30,248 images of 1041 persons, and its test set covers the remaining 3060 persons with 11,659 query images and 82,161 gallery images. For MSMT17\_V2, we split the training persons into 10 tasks also, each task with 104 persons (the first task with 105 persons).
\textbf{(3) VeRi-776~\cite{liu2016large}} contains 49,357 images of 776 vehicles, which are captured by 20 cameras. Among them, 576 vehicles are used for training and the remaining 200 are used for testing. In total, VeRi-776 consists of 37,778 training images, 1,678 query images, and 11,579 gallery images. For continual metric learning, we split the training 576 vehicles into 10 tasks, each task with 57 vehicles (the first task with 63 vehicles).

\minisection{The Lifelong ReID (LReID) benchmark.}
We adapt the train set of the inter-domain LReID benchmark by building it from four datasets: Market-1501~\cite{zheng2015scalable}, CUHK-SYSU ReID~\cite{xiao2016end}, MSMT17\_V2~\cite{wei2018person}, and CUHK03~\cite{li2014deepreid}.\footnote{We removed DukeMTMC-reID from the LReID benchmark due to its retraction on account of privacy issues.} After training, the model is evaluated on the test query and gallery sets LReID-Seen of these four datasets (i.e. over \textit{seen} domains). We also test on LReID-Unseen test set which combines seven person ReID datasets: VIPeR~\cite{gray2008viewpoint}, PRID~\cite{hirzer2011person}, GRID~\cite{loy2010time}, i-LIDS~\cite{zheng2009associating}, CUHK01~\cite{li2012human}, CUHK02~\cite{li2013locally}, and SenseReID~\cite{zhao2017spindle}.

\minisection{Implementation details.}
We follow the same network structure and training strategy as DMML~\cite{chen2019deep} for our method. The ResNet-50~\cite{he2016deep} pretrained on ImageNet~\cite{russakovsky2015imagenet} works as our feature extractor for all methods. The feature extractor is further trained during continual training. We use the Adam optimizer~\cite{kingma2014adam} with a base learning rate of $LR=0.0002$ and weight decay of $0.0001$. We set the trade-off coefficient to $\lambda=1.0$, the margin as $\tau=0.4$, and the temperature to $T=1.0$ for DwoPP and $T=10.0$ for DwPP. The number of classes, support and query images in each episode are $N=32, n_s=5, n_q=1$.

\minisection{Compared methods and metrics.}
Our evaluation is divided into two parts: 
    \textbf{(1)} To compare with conventional continual learning methods, we train models with the softmax-triplet loss of BoT~\cite{Luo_2019_CVPR_Workshops}. For methods using this loss without exemplars, we selected AKA~\cite{pu2021lifelong}, PASS~\cite{zhu2021prototype}, and LwF~\cite{li2017learning}. For methods using exemplars, we selected FT+, iCaRL~\cite{rebuffi2017icarl} and LwF+~\cite{li2017learning}.
    \textbf{(2)} For comparison with incremental meta learning methods, we build upon the DMML loss~\cite{chen2019deep}. For methods without exemplars we selected IDA~\cite{liu2020incremental}. For methods with exemplars, we selected ERD~\cite{wang2021incremental}.
Note that AKA is the state-of-the-art in LifelongReID and IDA is the state-of-the-art in incremental meta learning. For all exemplar-based methods we store 500 exemplars for all experiments. We use mean Average Precision (mAP) and Accuracy at Rank-1 as metrics~\cite{ye2021deep}. We compute the mAP and Rank-1 Accuracy of the model on the unseen test set after each task. All results are averages over three runs.

\begin{table*}
\begin{center}
\scalebox{0.63}{
\begin{tabular}{|r|cc|cc|cc|cc|cc|cc|}
\hline

\textbf{Metric:} & \multicolumn{6}{|c|}{\textbf{mAP}}  & \multicolumn{6}{|c|}{\textbf{Rank-1 Accuracy}}  \\
\hline

\textbf{Dataset:}& \multicolumn{2}{|c|}{{Market}}   & \multicolumn{2}{|c|}{{MSMT17}}  & \multicolumn{2}{|c|}{{{VeRi-776}}} & \multicolumn{2}{|c|}{{Market}}   & \multicolumn{2}{|c|}{{MSMT17}}  & \multicolumn{2}{|c|}{{VeRi-776}} \\
\hline\hline

\multicolumn{13}{|c|}{\textbf{Based on episodic optimization with DMML loss~\cite{chen2019deep}}}  \\
\hline
\textbf{Joint training:} &  \multicolumn{2}{|c|}{ 82.2} & \multicolumn{2}{|c|}{ 44.7} & \multicolumn{2}{|c|}{ 73.6} & \multicolumn{2}{|c|}{ 92.6} & \multicolumn{2}{|c|}{ 68.9} & \multicolumn{2}{|c|}{ 92.1}\\
\hline
\textbf{Sessions:} & last & avg & last & avg & last & avg & last & avg & last & avg & last & avg\\
\hline

\multicolumn{13}{|c|}{without exemplars}  \\
\hline

DMML-FT ({ICCV'19}) & 56.3 & 49.1 & 10.9 & 10.0 & 30.8 & 29.3 & 77.8 & 71.5 & 28.9 & 27.4 & 70.3 & 62.4 \\  

IDA ({ECCV'20})     & 32.2 & 37.8 & 19.2 & 16.8 & 21.0 & 18.4 & 58.7 & 63.1 & 45.6 & 38.2 & 56.6 & 45.4  \\ 

DwPP                & 57.8 & 48.4 & 16.3 & 13.3 & 30.9 & 28.9 & 78.1 & 70.7 & 39.0 & 33.9 & 71.7 & 63.3  \\ 

\textbf{\textit{Ours (DwoPP)}}       & \textbf{67.2} & \textbf{57.6} & \textbf{23.8}& \textbf{19.1} & \textbf{39.9} & \textbf{35.3} & \textbf{84.6} & \textbf{77.1}  &\textbf{51.0} &\textbf{42.6} & \textbf{78.5} & \textbf{69.3}  \\

\hline
\multicolumn{13}{|c|}{with 500 exemplars in total}  \\
\hline
ERD (CVPRW'22)               & 63.5 & 53.9 & 21.7 & 17.2 & 38.2 & 33.8 & 81.8 & 74.5 & 46.6 & 39.4 & 72.9 & 65.5 \\
\hline\hline
\multicolumn{13}{|c|}{\textbf{Based on global optimization with softmax-triplet loss from BoT~\cite{Luo_2019_CVPR_Workshops}}}  \\
\hline
\textbf{Joint training:} & \multicolumn{2}{|c|}{ 82.4 } & \multicolumn{2}{|c|}{ 43.2} & \multicolumn{2}{|c|}{ 69.2}  & \multicolumn{2}{|c|}{ 93.0} & \multicolumn{2}{|c|}{ 71.1 }  & \multicolumn{2}{|c|}{ 92.7} \\
\hline
\textbf{Sessions:} & last & avg & last & avg & last & avg & last & avg & last & avg & last & avg\\
\hline
\multicolumn{13}{|c|}{without exemplars}  \\
\hline
BoT-FT ({CVPR'19}) & 30.7 & 33.5 & 6.6  & 8.4  & 25.8 & 24.9 & 55.4 & 58.8 & 20.6 & 25.4 & 65.3 & 62.1  \\  
LwF ({ECCV'16})    & 40.5 & 40.2 & 10.7 & 11.8 & 31.2 & 28.1 & 65.9 & 65.4 & 30.3 & 32.3 & 71.3 & 65.8  \\
PASS ({CVPR'21})   & 40.0 & 40.1 & 9.9 & 11.6  & 30.7 & 27.3 & 65.8 & 64.2 & 29.7 & 31.9 & 70.9 & 64.4   \\
AKA ({CVPR'21})    & 52.5 & 45.6 & 15.1& 13.3  & 30.9 & 27.1 & 76.2 & 69.9 & 37.3 & 34.6 & 72.9 & 64.4  \\ 
\hline
\multicolumn{13}{|c|}{with 500 exemplars in total}  \\
\hline
BoT-FT+ ({CVPR'19}) & 61.5 & 52.4 & 21.5 & 17.5 & 36.7 & 32.3 & 81.0 & 74.4 & 47.7 & 41.3 & 76.2 & 69.6  \\ 
iCaRL ({CVPR'17})   & 58.0 & 52.2 & 21.6 & 18.3 & 38.0 & 33.3 & 78.7 & 74.5 & 47.5 & 42.2 & 78.1 & 70.9  \\ 
LwF+ ({ECCV'16})    & 60.7 & 54.0 & 20.8 & 17.5 & 38.3 & 33.3 & 80.3 & 75.4 & 46.6 & 40.8 & 77.9 & 70.1  \\

\hline

\end{tabular}
}
\end{center}
\caption{
Results in mAP and Rank-1 Accuracy (in \%) after last task and average over all tasks. The top half reports results for meta metric learning, and the lower half for global optimization methods using the softmax-triplet loss (BoT~\cite{Luo_2019_CVPR_Workshops}). Results are further split into methods with and without exemplars. The best exemplar-free results are highlighted in \textbf{bold}.
}
\label{tab:complete_table_3datasets}
\end{table*}

\subsection{Comparative performance evaluation}


\minisection{Intra-domain Lifelong Object ReID.} Fig.~\ref{fig:10_task_all_dataset} gives the mAP and Rank-1 curves on Market-1501, MSMT17\_V2, and VeRI-776. We report the performance of all methods after task $t=10$ and the average metrics over all training sessions in Table~\ref{tab:complete_table_3datasets}. On all three datasets, finetuning with softmax-triplet loss is always sub-optimal to finetuning with the meta metric loss. The performance gap between the  mAP for the DMML-FT and BoT-FT 
after the last task is $25.8$, $4.3$, and $5.0$ on three datasets, respectively. 
Note that the two losses result in a similar joint training performance. This demonstrates that meta metric learning is more suitable to the Continual Metric Learning problem, as we discussed in Sec.~\ref{sec:cml}. For continual learning methods without exemplars, our method DwoPP performs best on all datasets. Compared to the DMML-FT metrics after task 10, DwoPP improves by between $9.1$ to $12.9$ in mAP. 
Note that on Market-1501 and VeRi-776 DMML-FT outperforms most of the methods that actively counter fogetting.
Furthermore, we also include a comparison with rehearsal methods in Table~\ref{tab:complete_table_3datasets}. The methods iCaRL, LwF+, FT+ and ERD obtain similar results, and improved performance compared to DMML-FT.
Our exemplar-free method DwoPP performs better than exemplar-based methods on these three datasets (only marginally worse in average Rank-1 Accuracy on VeRi-776). We also ablate our distillation and report results for distillation with all pairs (DwPP). The results of DwPP show that naive application of knowledge distillation to continual meta metric learning does hardly improve results. The removal of positive pairs (DwoPP) results in large performance gains after the last task: gains between $7.5$ to $9.4$ in mAP.



\minisection{Inter-Domain Lifelong Person ReID (LReID).}
In Table~\ref{tab:lifelongreid}, we compare DwoPP with other methods on the LReID~\cite{pu2021lifelong} benchmark. Similar to the results for the intra-domain ReID setting, the DMML-FT baseline outperforms BoT-FT by a large margin for both seen and unseen tasks. Our method performs best, outperforming AKA by 8.5/9.6 (mAP/Rank-1 Accuracy) on seen tasks and 4.9/4.2 (mAP/Rank-1 Accuracy) on unseen tasks. The difference between DwPP and DwoPP on LReID further highlights the importance of removing positive pairs from knowledge distillation. See the Supplementary Material for more analysis.

\begin{table}[tb]
\centering
\scalebox{0.5}{
\begin{tabular}{|c|c|c|c|c|c|c||c|c|c|c|c|c|} 

\hline

& \multicolumn{6}{|c||}{\textbf{mAP}}  & \multicolumn{6}{|c|}{\textbf{Rank-1 Accuracy}}  \\
\hline
& \multicolumn{1}{|c|}{market} & \multicolumn{1}{|c|}{sysu} &\multicolumn{1}{|c|}{msmt17} &\multicolumn{1}{|c|}{cuhk03} & \multicolumn{1}{|c|}{\textit{\textbf{seen avg.}}} & \multicolumn{1}{|c||}{\textit{\textbf{unseen}}} & \multicolumn{1}{|c|}{market} & \multicolumn{1}{|c|}{sysu} &\multicolumn{1}{|c|}{msmt17} &\multicolumn{1}{|c|}{cuhk03} & \multicolumn{1}{|c|}{\textit{\textbf{seen avg.}}} & \multicolumn{1}{|c|}{\textit{\textbf{unseen}}}  \\ 

\hline\hline

BoT-FT  & 11.6 & 54.6 & 0.8 & 31.2 & 24.6 & 32.4 & 31.6 & 61.6 & 2.8 & 35.1 & 32.8 & 32.8  \\  
LwF &  21.0 & 58.0 & 1.7 & 48.0 & 32.2 & 43.3 &  46.5 & 64.7 & 5.8 & 53.8 & 42.7 & 42.9  \\ 
AKA   & 18.7 & 56.3 & 1.6 & 48.6 & 31.3 & 43.6  & 42.3 & 63.1 & 5.8 & 53.9 & 41.3 & 43.6  \\ 

\hline

DMML-FT & 22.5 & 56.8& 2.3 & 67.0 & 37.2 & 42.8 & 47.3 & 62.6 & 8.4 & \textbf{73.8} & 48.0 & 42.6	\\ 
DwPP &23.2 &	56.7 &	2.2	 & \textbf{67.9} &	37.5 &	44.7 & 49.1 & 	63.2 & 	7.5	 & 72.4	 & 48.0	 & 44.2 \\
Ours (DwoPP) &   \textbf{34.4} &  \textbf{67.3} &  \textbf{4.1} &  {53.5} &  \textbf{39.8 }&  \textbf{48.5}  &   \textbf{58.6} &\textbf{73.0} &\textbf{12.3} &59.6 &\textbf{50.9} &\textbf{47.8} \\
\hline

\end{tabular}
}
\caption{Results after learning the last task.
BoT~\cite{Luo_2019_CVPR_Workshops} (above) and DMML~\cite{chen2019deep} (below).
}
\label{tab:lifelongreid}
\end{table}

\begin{figure*}
\begin{centering}
\begin{tabular}{ccc}
\includegraphics[width=0.3\textwidth]{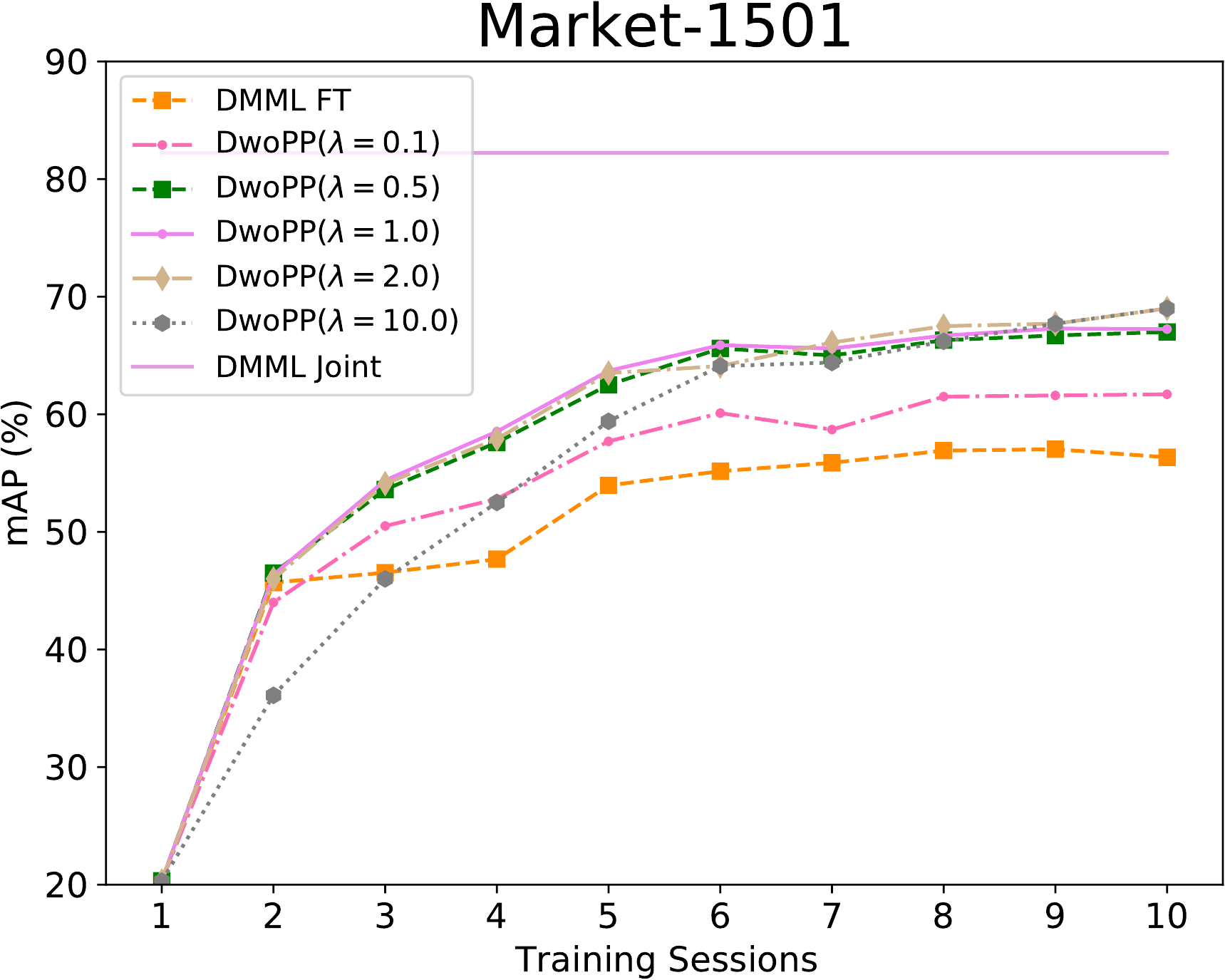} &
\includegraphics[width=0.3\textwidth]{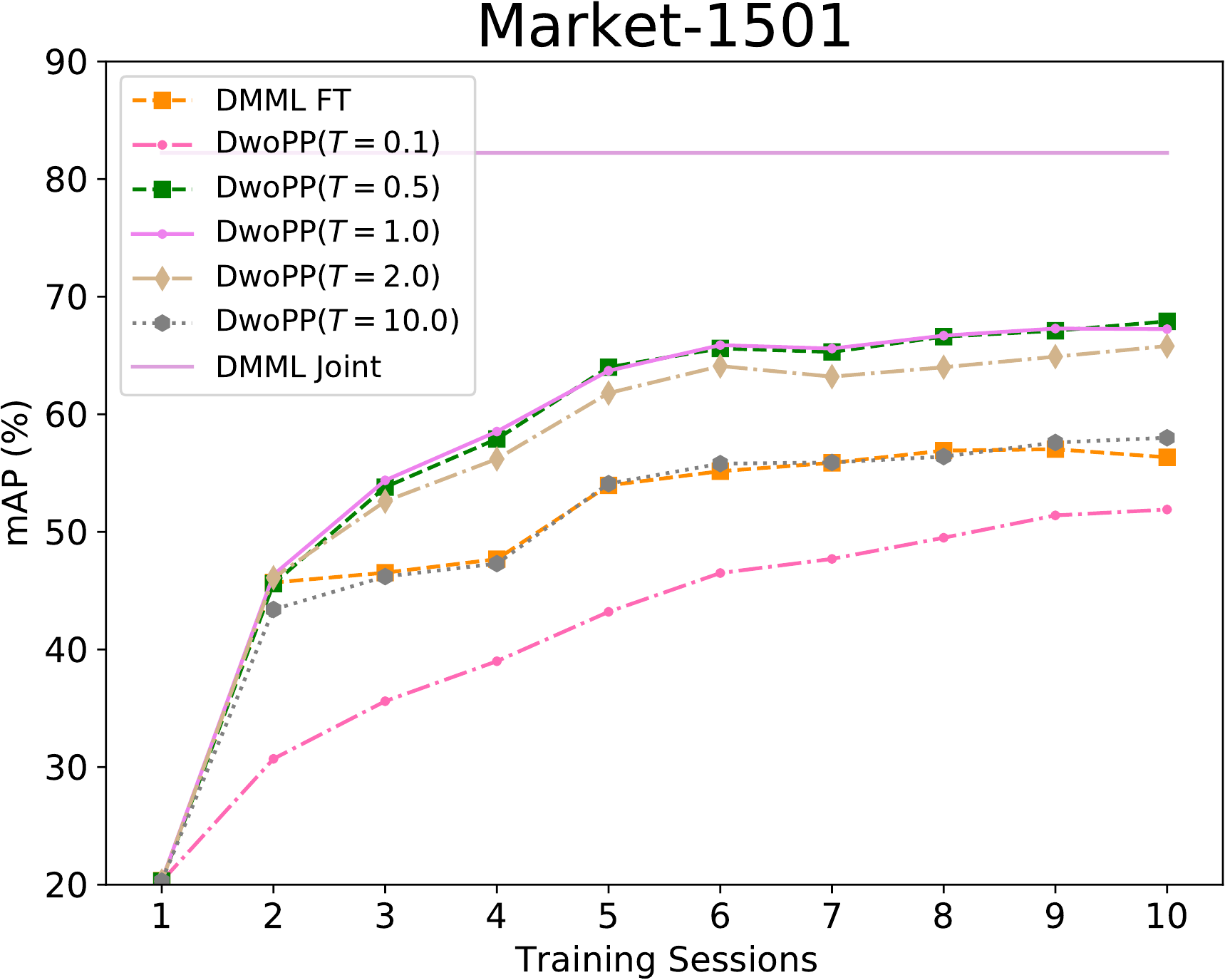} &
\includegraphics[width=0.3\textwidth]{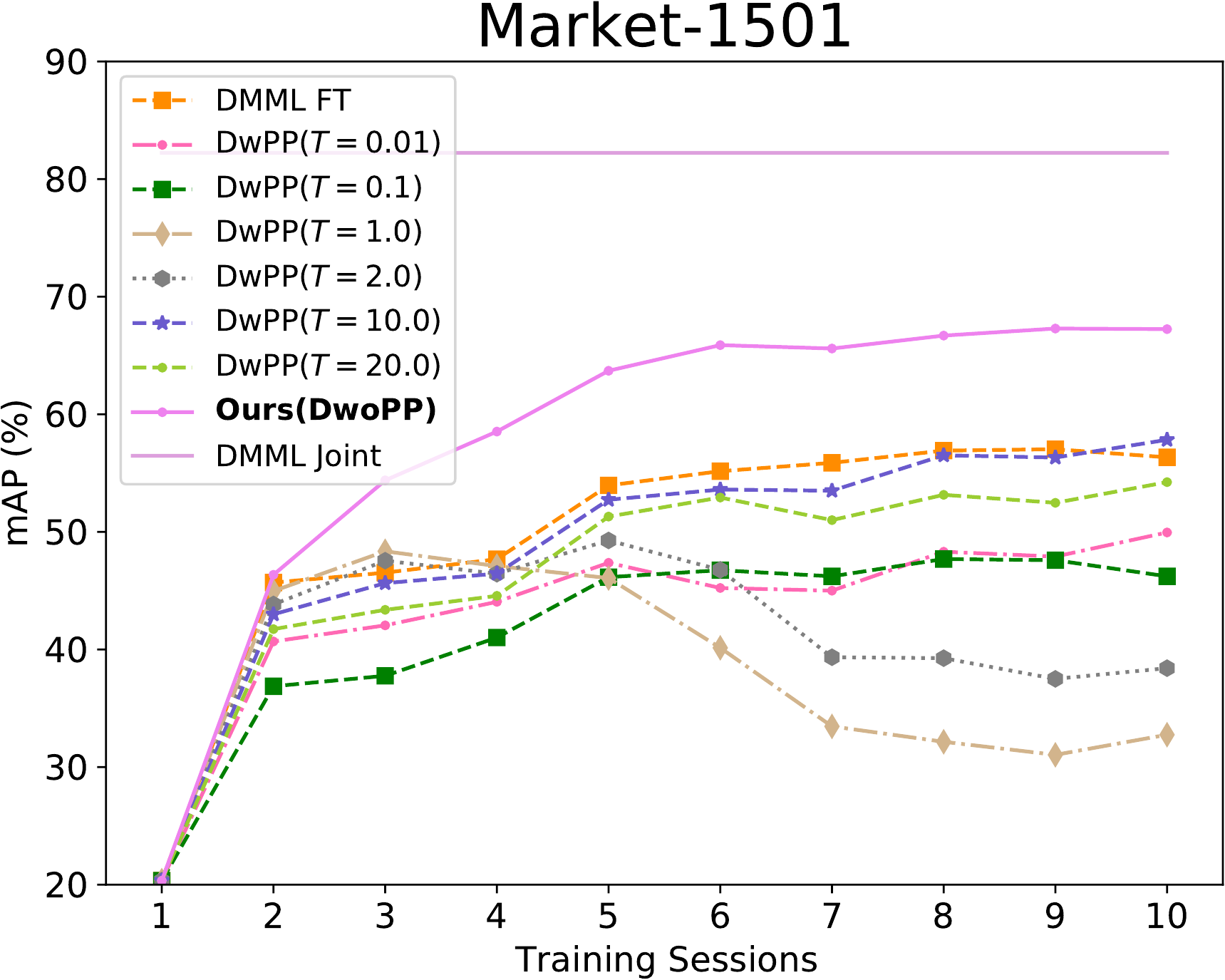} \\
(a) $\lambda$ for DwoPP & (b) $T$ for DwoPP & (c) $T$ for DwPP
\end{tabular}
\caption{Ablation study on hyperparameters $\lambda$ and $T$.}
\label{fig:ablations}
\end{centering}
\end{figure*}

\minisection{Influence of positive pairs on distillation.}
To better understand the role of positive pairs (PP) and negative pairs (NP) in knowledge distillation, we decouple the knowledge distillation (following DKD~\cite{zhao2022decoupled}) from Eq.~\ref{eq:DwPP_distillation} into PPKD and NPKD by $\mathcal{L}_{\text{DwPP*}}=\alpha* PPKD + \beta* NPKD, \alpha+\beta=1.0$ (here we use $T=1.0$). Note that  $\mathcal{L}_{\text{DwPP}}= PPKD + \rho* NPKD$ (see Supplementary Material for further explanations).
In Table~\ref{tab:decoupled_kd}, we observe that the  performance drastically decreases with higher participation of positive pairs.


\minisection{Ablation on $\lambda$ in DwoPP and temperature $T$ in both DwoPP and DwPP.} In Fig.~\ref{fig:ablations}(a) we vary $\lambda$ which controls the tradeoff between metric and distillation losses. Except for $\lambda=10.0$ and $\lambda=0.1$, DwoPP performance is stable to changing $\lambda$. We set $\lambda=1.0$ for DwoPP in all experiments.
In Fig.~\ref{fig:ablations}(b), we vary the temperature hyperparameter $T$ in DwoPP. A high temperature smooths the distribution and decreases the influence of the dominant class. For DwoPP $T=10.0$ performs similarly to finetuning, and $T=0.1$ causes the model to focus only on the highest probability. Thus we set $T=1.0$ for DwoPP. In Fig.~\ref{fig:ablations}(c) we vary the temperature $T$ in DwPP to determine if larger temperatures benefit it. However, even with the best $T=10.0$, DwPP performs similarly to DMML-FT and much worse than DwoPP. Again showing that naive knowledge distillation does not improve results for continual meta metric learning. We use $T=10$ for DwPP in all experiments.

\begin{table}
\centering
\scalebox{0.75}{
\begin{tabular}{|c|c|c|c|c|c|c|c|} 
\hline
  & & DwoPP  & \multicolumn{4}{|c|}{DKD~\cite{zhao2022decoupled} } & DwPP   \\ 
\hline
 & $\alpha$ & 0.0   & 0.1  & 0.3  & 0.5  & 1.0  & 1.0 \\

 & $\beta$ & 1.0   & 0.9  & 0.7  & 0.5  & 0.0  & 1-$\rho$ \\
\hline
\multirow{2}{*}{mAP}   & last  & \textbf{67.2}  & 62.9 & 48.2 & 36.0 & 25.9  & 32.8 \\ 
\cline{2-8}
 & avg   & \textbf{57.6}  & 53.7 & 46.8 & 39.1 & 32.1& 37.8 \\  

\hline

\end{tabular}
}
\caption{
Decoupling Eq.~\ref{eq:DwPP_distillation} into PPKD and NPKD with coefficients $\alpha$ and $\beta$ on Market-1501 with temperature $T=1.0$. $\rho$ is the positive probabilities as in DKD~\cite{zhao2022decoupled}.
\vspace{-0.1in}
}
\label{tab:decoupled_kd}
\end{table}


\section{Conclusions}
We demonstrate that meta learning approaches perform better than those based on global metric loss optimization for Object ReID. We therefore proposed an approach based on Continual Meta Metric Learning. 
To overcome forgetting, we propose Distillation without Positive Pairs (DwoPP) as an approach that eliminates positive samples from distillation. This distillation makes the metric learning model accumulate knowledge from the previous and current tasks and generalize better to unseen tasks. Extensive experiments on newly proposed intra-task object re-identification datasets and the existing LReID benchmark demonstrate the effectiveness of our approach. Furthermore, experiments confirm that naive knowledge distillation does not improve results for continual meta metric learning, and only after the removal of positive pairs is forgetting of previous tasks effectively countered.

\section*{Acknowledgement}
We acknowledge the support from Huawei Kirin Solution, the Spanish Government funded project PID2019-104174GB-I00/AEI/10.13039/501100011033., and from the European Commission under the Horizon 2020 Programme, grant number 951911 – AI4Media.

\bibliography{longstrings,mybib}



\section*{Supplementary material}

# Supplementary Material:
# Positive Pair Distillation Considered Harmful: Continual Meta Metric Learning for Lifelong Object Re-Identification

Kai Wang[1]*
kwang@cvc.uab.es

Chenshen Wu[1]*
chenshen@cvc.uab.es

Andy Bagdanov[3]
andrew.bagdanov@unifi.it

Xialei Liu (*corresponding author*)[2]
xialei@nankai.edu.cn

Shiqi Yang[1]
syang@cvc.uab.es

Shangling Jui[4]
jui.shangling@huawei.com

Joost van de Weijer[1]
joost@cvc.uab.es

[1] Computer Vision Center
Universitat Autònoma de Barcelona
Barcelona, Spain

[2] College of Computer Science
Nankai University
Tianjin, China

[3] MICC
University of Florence
Florence, Italy

[4] Huawei Kirin Solution
Shanghai, China

## 1    Implementation details.

We follow the same network structure and training strategy as DMML [1] for methods based on the DMML loss, and use the network and training protocol of BoT [10] for methods based on the softmax-triplet. For our person ReID experiments, we use ResNet-50 [3] pretrained on ImageNet [12] as our feature extractor. The last spatial downsampling operation in the network is removed to maintain high resolution. We resize input images to $256 \times 128$ for all methods. For vehicle ReID, we also use a ResNet-50 backbone pretrained on ImageNet as the embedding architecture, and use input images of size $224 \times 224$ augmented with random horizontal flips. We use the Adam optimizer [5] with a base learning rate of $LR = 0.0002$ and weight decay of 0.0001. All models are trained for 600 epochs with fixed learning rate of 0.0002 for the first 300 epochs, after which the learning rate is reduced by a factor of $0.005^{1/300}$ each epoch until the end. We set the trade-off coefficient to $\lambda = 1.0$, the margin as $\tau = 0.4$ as in DMML [1], and the temperature to $T = 1.0$ for DwoPP and $T = 10.0$ for DwPP. The number of classes, support images, and query images are in each episode are $N = 32, n_s = 5, n_q = 1$, respectively.

| Continual learning metrics on Market 1501 | | | | | | | |
|---|---|---|---|---|---|---|---|
| | BoT FT | LwF | AKA | IDA | DMML FT | DwPP | DwoPP |
| Plasticity | 9.7 | 9.5 | 9.7 | 7.2 | 10.8 | 10.7 | 8.1 |
| Forgetting | -8.9 | -7.7 | -6.5 | -5.9 | -6.8 | -6.5 | -2.9 |
| Overall | 0.8 | 1.8 | 3.2 | 1.3 | 4.0 | 4.2 | 5.2 |

Table 4: Average forgetting and plasticity in mAP (%) on Market-1501 together with the overall mAP change (defined as plasticity plus forgetting).

| | | continual metric learning training tasks | | | | Unseen-test task |
|---|---|---|---|---|---|---|
| *Task id:* | | 1 | 2 | ... | 10 | |
| | Market-1501 | 76 | 75 | ... | 75 | 750 |
| Identities per task: | MSMT17_V2 | 105 | 104 | ... | 104 | 3060 |
| | VeRi-776 | 63 | 57 | ... | 57 | 200 |

Table 5: Our proposed 10-task split of two Person ReID datasets and one Vehicle ReID dataset for Continual Metric Learning.

# 2 Evaluating forgetting.

To further analyze the results we measure forgetting and plasticity on the Market 1501 dataset. Continual learning aims to counter forgetting (stability) while optimally learning new tasks (plasticity). To measure these, we track the change in mAP for each identity in the unseen test set after each task: a drop is added to forgetting, an increase to plasticity. In Table 4 we report the plasticity and forgetting averaged over tasks. We see that DwoPP has greatly reduced forgetting at the price of only a small decrease in plasticity.

# 3 Continual Metric Learning splitting protocols

Our proposed Continual Metric Learning splits for two Person ReID datasets and one Vehicle ReID dataset are shown in Table. 5. For all three datasets, we try to uniformly distribute the training identities into 10 continual metric learning tasks. The query and gallery set are fixed and serve as the unseen-test task.

# 4 DMML loss illustration

An illustration of the DMML loss [1] is shown in Fig. 5. The hard-mining DMML loss finds the largest distance to a positive example and the smallest distance to a negative sample to compute the metric loss with a margin.

# 5 More random orders on Market-1501 dataset.

In the main paper, we split tasks according to the *object IDs*. Thus, for the purpose of verifying the robustness of our proposed method to various orderings of the tasks, we randomly generate three different orderings of person IDs from Market-1501 to split the tasks (see Fig. 6). The results show that the trends are similar as those reported in Table 1. Results are averages and standard deviations in mAP and Rank-1 Accuracy over these three runs.

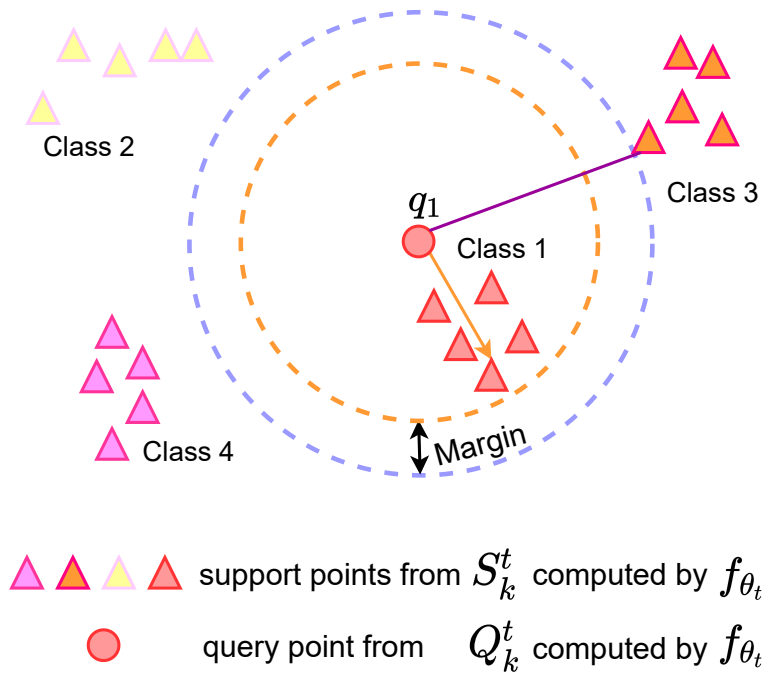

Figure 5: Supposing the query point is from class 1, the hard-mining DMML loss selects the farthest positive point and nearest negative point to compute the distance. It forces a margin between negative and positive distances.

# 6   Lifelong Person ReID (LReID) benchmark

For the LReID benchmark[11] we removed the DukeMTMC-ReID dataset due to its retraction on account of privacy issues. Except for this change, we keep the same training order as LReID Order-1: Market-1501 [16]→ CUHK-SYSU [14] → MSMT17_V2 [13] → CUHK03 [8]. After training each task, we evaluate the model over the test query and gallery sets in LReID-Seen for these four datasets, and also on the LReID-Unseen test set consisting of seven person ReID datasets: VIPeR [2], PRID [4], GRID [9], i-LIDS [17], CUHK01 [7], CUHK02 [6], and SenseReID [15]. All the performance curves on LReID-Seen are shown in Fig. 8 and the curves on LReID-Unseen are shown in Fig. 7.

An interesting phenomenon we observed in the main paper is that DwPP is always better in the current task evaluation. We assume this is because DwPP forces the predictions to be aligned with the probability distributions of the old model, which contain some information about the relative distances of these identities. This extra information further enhances representation learning in the current task, thus leading to better performance on the current task even compared to the finetuning baseline (which is usually better on the current task).

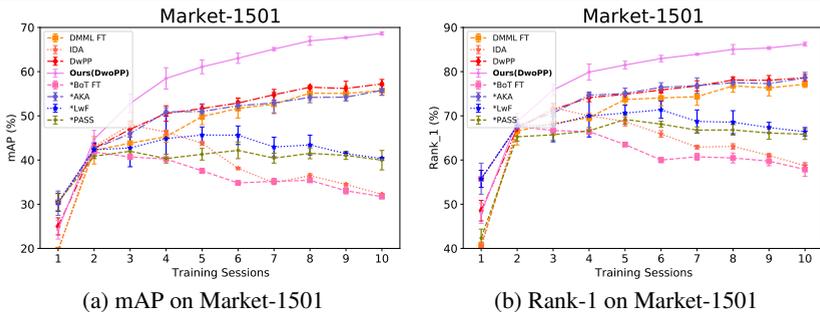

(a) mAP on Market-1501　　　　　　(b) Rank-1 on Market-1501

Figure 6: Performance on Market-1501 averaged over three random ID orders with standard deviation.

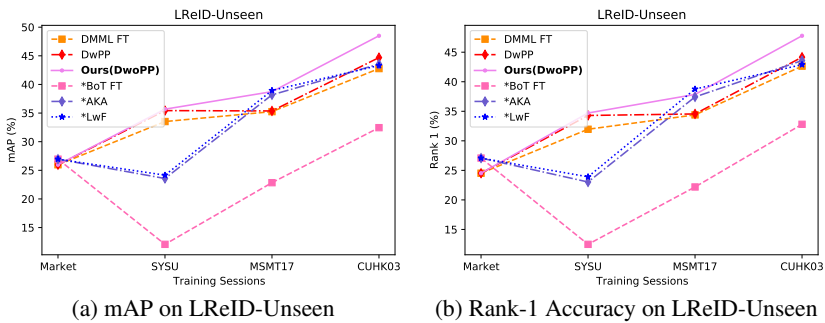

(a) mAP on LReID-Unseen　　　　　　(b) Rank-1 Accuracy on LReID-Unseen

Figure 7: Results in mAP and Rank-1 Accuracy the LReID-Unseen test set of the LReID benchmark. The training order is (Market-1501→ CUHK-SYSU → MSMT17_V2 → CUHK03).

# 7 Limitations and ethical considerations

Person ReID is fraught with ethical concerns over its potential to violate the privacy of observed subjects. Although continual learning for Person ReID offers the possibility of learning and updating models without the need for long-term retention of sensitive data, it also runs the risk of "baking" biases into the model that, due to mitigation of forgetting, become difficult to remove. For real applications there is still a large gap between joint and continual training for object ReID, and a limitation of the experiments in this work is the relatively short task sequences we consider.

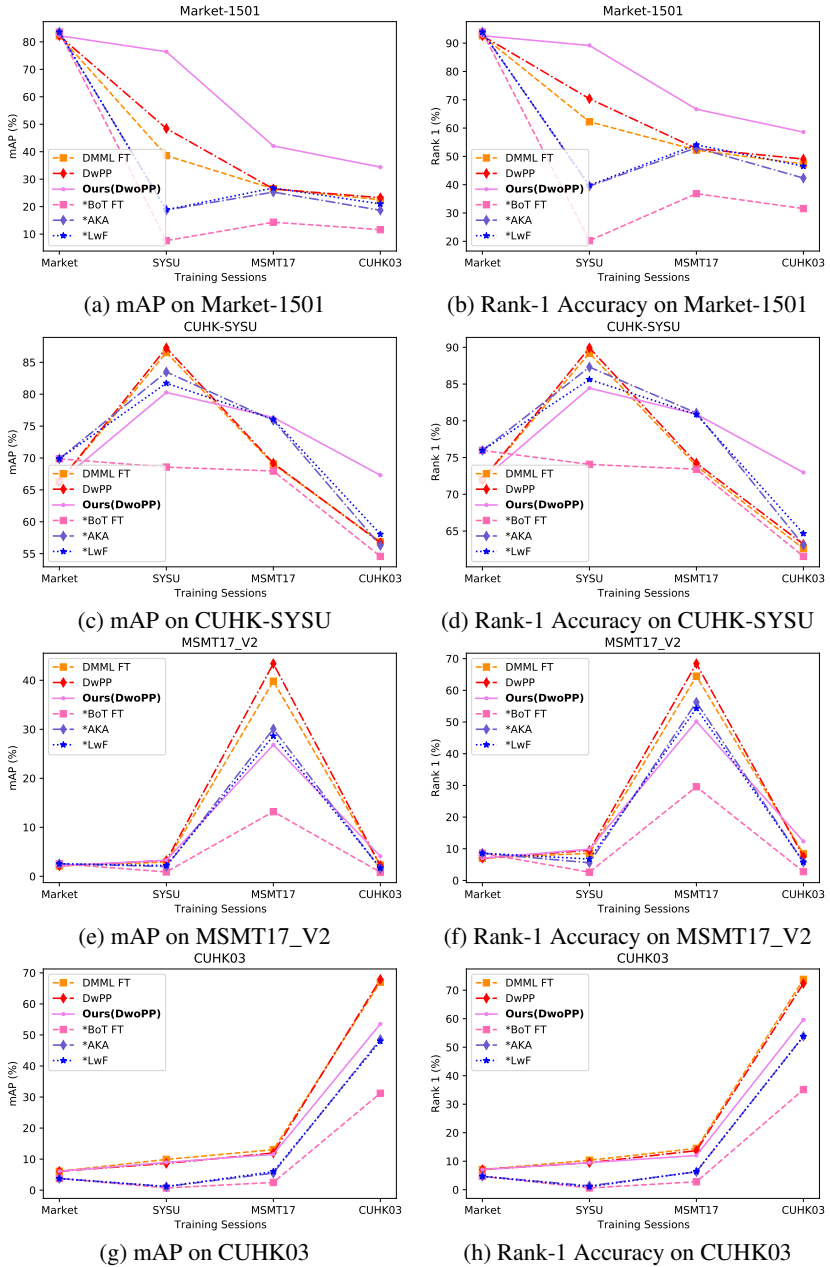

Figure 8: Results in mAP and Rank-1 Accuracy on the LReID benchmark. The training order is (Market-1501→ CUHK-SYSU → MSMT17_V2 → CUHK03). The first four rows show the evaluation on these four tasks respectively.



\end{document}